# Personalized Prediction of Perceived Message Effectiveness Using Large Language Model–Based Digital Twins

Jasmin Han[1], Janardan Devkota[1], Joseph Waring[1], Amanda Luken[2], Felix Naughton[3], Roger Vilardaga[4], Jonathan Bricker[5,6], Carl Latkin[7], Meghan Moran[7], Yiqun Chen[8,9,*], Johannes Thrul[1,10,11*]

1 Department of Mental Health, Johns Hopkins Bloomberg School of Public Health, Baltimore, MD, USA

2 Department of Health Sciences, Towson University, Towson, USA

3 Addiction Research Group, University of East Anglia, Norwich, UK

4 Department of Implementation Science, Wake Forest University School

of Medicine, Winston-Salem, USA

5 Fred Hutchinson Cancer Center, Seattle, USA

6 Department of Psychology, University of Washington, Seattle, USA

7 Department of Health, Behavior and Society, Johns Hopkins Bloomberg

School of Public Health, Baltimore, USA

8 Department of Biostatistics, Johns Hopkins Bloomberg

School of Public Health, Baltimore, USA

9 Department of Computer Science, Johns Hopkins Whiting School of Engineering, Baltimore, USA

10 Sidney Kimmel Comprehensive Cancer Center at Johns Hopkins, Baltimore, USA

11 Centre for Alcohol Policy Research, La Trobe University, Melbourne, Australia

* Drs. Chen and Thrul share last authorship

**Corresponding author**: Jasmin Han, Department of Mental Health, Johns Hopkins Bloomberg School of Public Health, Baltimore, MD, jhan94@jhu.edu

**Abstract (250/250)**

**Objective:** Perceived message effectiveness (PME) by potential intervention end-users is important for selecting and optimizing personalized smoking cessation intervention messages for mobile health (mHealth) platform delivery. This study evaluates whether large language models (LLMs) can accurately predict PME for smoking cessation messages.

**Materials and Methods:** We evaluated multiple models for predicting PME of smoking cessation messages across three domains: content quality, coping support, and quitting support. The dataset comprised 3,010 message ratings (5-point Likert-scale) from 301 young adult smokers. We compared (1) supervised learning models trained on labeled data, (2) zero-/few-shot LLMs prompted without task-specific fine-tuning, and (3) LLM-based digital twins, which incorporate individual characteristics and prior PME histories to generate personalized predictions. Model performance was assessed on three held-out (of ten) messages per participant using accuracy, Cohen's κ, and F1.

**Results:** LLM-based digital twins outperformed zero-/few-shot (+12 percentage points on average) LLMs and supervised baselines (+13 percentage points), achieving accuracies of 0.49 (content), 0.45 (coping), and 0.49 (quitting), with corresponding directional accuracies of 0.75, 0.66, and 0.70 for simplified 3-point scale. Digital twin predictions also showed greater dispersion across rating categories, indicating improved sensitivity to individual differences.

**Discussion:** We found that integrating personal profiles with LLMs captures person-specific differences in PME and outperforms supervised learning and zero-/few-shot LLM approaches. This improved PME prediction could enable more tailored intervention content in mHealth.

**Conclusion:** LLM-based digital twin models show potential for predicting PME and may support personalization of mobile smoking cessation and other substance use and health behavior change interventions.

**Key words:** Large language models (LLMs); digital twins; perceived message effectiveness (PME); smoking cessation; personalized interventions

**Introduction**

Despite the decline in US smoking prevalence (Centers for Disease Control and Prevention, 2024; Meza et al., 2023), combustible tobacco use remains a major health issue among young adults. Young adulthood is a critical developmental period for smoking initiation (Villanti et al., 2015) and the formation of long-term nicotine dependence trajectories (U.S. Department of Health and Human Services, 2017). With the rapid expansion of wearable sensors and smartphone-based data collection, flexible and low burden mobile health (mHealth) interventions, especially text-based interventions, have demonstrated strong feasibility, acceptability, and initial efficacy in promoting smoking cessation (S. Li et al., 2025; Zhou et al., 2023).

A central principle underlying the effectiveness of mHealth interventions in recent years is personalization (Casu et al., 2025; Kashefi et al., 2024). Personalized smoking cessation programs aim to move beyond one-size-fits-all messaging by using individual-level information, such as demographic characteristics, smoking history, motivational states, psychological profiles, and momentary behavioral patterns (Businelle et al., 2022; Hébert et al., 2025; Li et al., 2024; Lin et al., 2023; Luken et al., 2023; Thrul et al., 2025), to tailor message content, timing, and intensity.

As a result, understanding and predicting individual-specific perceptions of message effectiveness are important tasks in developing evidence-based, personalized mHealth interventions, because perceived message effectiveness (PME) is associated with downstream changes in beliefs, intentions, and even smoking cessation behavior (Noar et al., 2020) and is therefore pivotal for intervention optimization and trial design. Accordingly, much formative work in mHealth has relied on user-centered qualitative methods, including focus groups, in-depth interviews, and iterative co-design, to identify which features, message framings, and interaction patterns users like or dislike, or adding additional rounds to refine prototypes across multiple cycles (Abroms et al., 2015; Jamison et al., 2013; Nagawa et al., 2022; Tucker et al., 2020). Building on qualitative work, quantitative techniques, including surveys and experimental designs have also been used to understand PME. Despite providing valuable insights for message development, these approaches add considerable lead-in time to the development and deployment of mHealth studies. Supervised learning models, including regression and tree-based models, have been widely used to analyze how factors such as message attributes, demographics, smoking behaviors, and behavioral-economic traits are related to PME (Hamoud et al., 2025; Solnick et al., 2021; Thrasher et al., 2018; Tripp et al., 2021). However, most of this work is explanatory in nature, focusing on estimating correlational relationships or assessing variable importance rather than generating individual-level predictions of PME.

Moving from explanation to prediction is critical for intervention design, as personalized message selection requires reliable prediction of how a specific individual is likely to respond to a given message. With the rise of artificial intelligence (AI), recent years have witnessed a rapid expansion of studies adapting off-the-shelf large language models (LLMs) to perform predictive tasks. One growing line of work examines the use of LLMs in zero-shot and few-shot learning settings (Brown et al., 2020; Li, 2023), where models operate with little or no prior examples with known outcomes from the target study population. These approaches have gained increasing attention in health-related domains (Fernández-Pichel et al., 2024; Jin et al., 2025; Labrak et al., 2024), where data annotation is resource-intensive and often limited in scale. Emerging evidence suggests that, even with minimal task-specific supervision, LLMs can support a range of downstream health applications by transferring general domain knowledge acquired during large-scale pretraining (Agrawal et al., 2022; Ge et al., 2023; Neves et al., 2025; Singhal et al., 2025). For example, Singhal et al. (2025) used this approach to generate answers to medical questions. Among these health applications are classification tasks, which are also the tasks addressed in this paper. On the other hand, another line of research has focused on personalization and explored persona-based prompting and detailed contextual backstories to condition large language model outputs on individual attributes. This work is referred to as digital twin (Katsoulakis et al., 2024) approaches. Digital twins use individualized conditioning, such as prior decisions, clinical histories, behavioral trajectories, or psychometric profiles, to condition LLM outputs to represent a specific person (Chen et al., 2025; R. Li et al., 2025; Makarov et al., 2025; Sprint et al., 2024; Toubia et al., 2025). By conditioning on real-world data, digital twins aim to approximate not only typical human behavior but the idiosyncratic decision-making patterns of a particular individual, and could be useful to assist in development and refinement of health behavior change intervention approaches. For instance, Sprint et al. (2024) used this approach to predict patient cognitive health diagnosis. Taken together, developments in LLMs and LLM-based digital twins have the potential to dramatically reduce the time and financial resources required for early-stage hypothesis testing, intervention refinement, and product development, which is particularly valuable for mHealth personalized interventions such as those designed for smoking cessation.

In this study, we analyzed data from a panel of young adult smokers who evaluated smoking cessation messages. Using these data, we systematically benchmarked a broad suite of LLMs and prompting techniques to predict PME at the individual level. Our contributions are twofold: (1) we explore predictive modeling of PME by evaluating seven different prompting strategies across five LLMs, and (2) we introduce a pilot analytic framework that generalizes to intervention-evaluation settings within and beyond smoking cessation, particularly under realistic personalization conditions characterized by limited per-person data.

## 2. Methods

### 2.1 Study design and population

We conducted a secondary analysis of data from an online panel study in which young adult smokers evaluated smoking cessation messages (Hamoud et al., 2025). The parent study was designed to assess the effectiveness of evidence-based digital health interventions grounded in Cognitive Behavioral Therapy (CBT) (Perkins et al., 2013) and Acceptance and Commitment Therapy (ACT) (Kwan et al., 2024) frameworks for promoting smoking cessation among diverse young adults: CBT-based messages emphasize strategies for managing cravings by redirecting attention through specific tasks or behaviors that serve as distractions. In contrast, ACT-based messages encourage individuals to acknowledge and accept cravings without judgment while maintaining focus on the present moment. Both therapeutic approaches have been extensively investigated in prior behavioral intervention research and provide empirically supported foundations for digital cessation programs.

The study population consisted of 301 young adults aged 18 to 30 years, recruited through an online Qualtrics research panel. Eligible participants had smoked at least 100 cigarettes in their lifetime, reported smoking every day or on some days, and were either currently attempting to quit or intending to quit within the next month. During the trial, participants evaluated 124 smoking cessation messages that were developed by the research team based on evidence from prior studies and paired with image content obtained from free stock photo websites (Pexels and Unsplash). Each participant rated 10 randomly selected messages, including five based on CBT strategies and five based on ACT strategies, resulting in a total of 3,010 message ratings or an average of 24 ratings per message (Kim & Cappella, 2019).

Participants completed an online survey in which they rated each message across four domains: perceived quality of *content* ("How would you rate the content (that is, the words and meaning) of this message?"); (2) perceived quality of *design* ("How would you rate the design (that is, how the message looks) of this message?"); (3) perceived message support for *coping* with smoking urges ("How helpful would this message be to support you in coping with a smoking urge or craving?"); (4) perceived message support for *quitting* smoking ("How helpful would this message be to support you in quitting or reducing smoking?"). Responses for content and design were rated on a five-point Likert scale with the following options: Very poor, Poor, Acceptable, Good, and Very good. Responses for coping support and quitting support were rated on a five-point Likert scale with the following options: Not at all helpful, Somewhat helpful, Moderately helpful, Very helpful, and Extremely helpful. Participants were also given the option to provide written feedback on the messages. In addition to message ratings, participants completed questions assessing their sociodemographic characteristics (age, sex, sexual orientation, race/ethnicity, household income, and education level), smoking behaviors and cessation factors (number of days smoked in the past 30 days, average cigarettes per smoking day, living with smokers, having friends who smoke, past-year quit attempts, motivation to quit, social support for quitting, current daily smoking status, time to first cigarette, and quit intention), and psychological flexibility, which was measured using the 7-item Acceptance and Action Questionnaire–II (AAQ-II(Bond et al., 2011)).

**2.2 Models**

To explore how LLMs can support the design of more effective smoking cessation interventions, we implemented a series of models to predict multiple dimensions of smoking cessation message ratings. In this paper, we focus on three of the four rating dimensions, i.e., content, coping, and quitting, since design reflects the characteristics of the accompanying image. We excluded the design dimension for both methodological and practical considerations. Preliminary analysis revealed high correlations between design and content ratings, indicating that participants evaluated messages holistically rather than distinctly separating visual design elements from textual content. Additionally, given that current LLMs (October 2025) perform markedly better on text than image inputs, focusing our analysis on the three text-based ratings could improve modeling efficiency while maintaining the core predictive capabilities needed to assess PME, aligned with our primary interest in optimizing the text content of intervention messages.

We evaluated methods across three categories: (1) traditional supervised learning baselines, (2) LLMs with zero-/few-shot settings, and (3) "digital twin" approaches, implemented as LLMs conditioned on (i) structured participant profiles and (ii) participant-specific histories of prior messages and ratings. For the LLMs and digital twin categories, we evaluated five popular LLMs with default API parameters: GPT-4o-mini and GPT-5 (OpenAI), DeepSeek-R1 (DeepSeek), Grok-4-Fast (xAI), and Gemini-2.5-Pro (Google).

For all the models, we used within-participant splits (7 messages for training/history; 3 held out for testing). Supervised baselines were fit on the 7 labeled messages and evaluated on the 3 held-out messages. Zero-/few-shot LLMs received only the held-out message text (plus instructions/demonstrations). Digital-twin LLMs additionally received the participant profile and the 7 history message–rating pairs, but never the held-out messages or ratings. Prompts were programmatically assembled from message IDs to ensure held-out items could not be included in the history block; results are reported on held-out messages unless noted.

**2.2.1 Supervised learning models based on patient characteristics only**

Two conventional machine learning models served as non-LLM baselines: a regularized logistic regression (L2 penalty with C=1.0) and a random forest classifiers (default parameters in scikit-learn; Kramer, 2016). Both models utilized participant-level features only, including age, gender, race/ethnicity, nicotine dependence (time to first cigarette), cigarettes per day, and psychological flexibility. Model training and evaluation followed the same data splits applied to the LLM-based methods described above.

**2.2.2 Zero-shot and few-shot LLMs**

For LLMs, we compared five prompt strategies that varied in the amount of participant context and calibration structure provided to the model. We include the detailed prompts in Appendix A1.

1. Zero-shot with all features (*Zero-shot (all)*): This template included all individual characteristics (sociodemographic variables, smoking behaviors and cessation factors, and psychological flexibility items, totaling 23 features) and queried the LLMs to generate categorical predictions for content, coping, and quitting.

2. Zero-shot with selected features (Zero-shot (select)): Here, we restricted the metadata to five high-coverage variables associated with smoking patterns and cessation readiness: Age, sex, race/ethnicity, self-reported motivation to quit, and perceived social support. These variables were chosen based on the magnitude of correlation. The analysis was designed to test whether a small yet informative set of individual characteristics could lead to performant LLM predictions compared to 1.

3. Few-shot with all features (*Few-shot (all)*): This template extended the zero-shot prompt in 1. by prepending two randomly-selected examples with extreme ratings per domain ("Very good"/"Extremely helpful" and "Very poor"/"Not at all helpful") sampled from the training set. Each example included participants' characteristics, message content, and ratings. This setup allows the model to observe how different messages are evaluated by participants with different profiles across the full range of ratings, before generating its predictions.

4. Few-shot with selected features (*Few-shot (select)*): This template mirrored the configuration described in 3 but truncated participant features in both the example and the target query to the five selected characteristics. It was designed to test whether reference examples continue to improve performance when the prompt is aggressively compressed.

5. Zero-shot with selected features and probability-like outputs: (Zero-shot (w/ prob)): This variant used the same configuration as the zero-shot with selected, while prompting the LLM to provide relative likelihood scores for each rating category. Although these scores are directly reported by the LLM and do not represent calibrated probabilities, they are treated as probability-like outputs to support the final classification.

### 2.2.3 LLM-based digital twins

This framework can be viewed as an extension of the few-shot LLM paradigm, with the few-shot examples now contextualized at the individual level, incorporating each participant's persona and 7 historical responses (Toubia et al., 2025) to build participant profiles. Using these profiles, we evaluated two configurations: (1) a basic profile with all individual characteristics, and (2) an enhanced profile that also incorporated random-forest predictions as prior information. An example prompt for the digital twins is provided in Appendix A1.

### 2.3 Evaluation metrics

We evaluated model performance using five metrics. First, we used exact accuracy, defined as the proportion of predictions that matched the true message rating on the five-point Likert scale. This metric is straightforward to interpret but might be non-discriminative in the presence of class imbalance, e.g., our dataset skewed toward the two most positive categories ("Very good/Extremely helpful" and "Good/Very helpful"). Moreover, accuracy does not differentiate between types of prediction errors (e.g., over- versus under-prediction of the scores). Second, we used Cohen's κ, which quantifies the model-vs.-ground truth agreement after adjusting for chance. Third, we included macro-averaged F1 across the five Likert scales. F1 score balances precision (how many predicted positives were correct) and recall (how many actual positives were detected), and the macro-average allows each class to contribute equally to the overall performance, but might over-penalize errors on rare classes.

While accurately predicting the granular five classes was our primary goal, practical message optimization often only needs the ratings to be directional correct. We therefore collapsed ratings into three categories, mapping Very helpful and Extremely helpful to positive, Somewhat helpful and Not at all helpful to negative, and retaining Neutral. Based on these categories, we computed (1) directional accuracy, which captures whether predictions align with the correct direction of change, and (2) directional macro-F1, computed on directional agreement rather than exact rating match. We also report bootstrap confidence intervals for the metrics in Appendix A2, complementing the point estimates in the main text.

## 2.4 Top-K Message Selection Evaluation

From a translational perspective, investigators seek to leverage these models to prioritize intervention content for real-world deployment. To evaluate the practical utility of LLM-based message scoring, we assessed whether Digital Twin prompting enables efficient identification of highly rated messages. We treated ordinal rating scales as numeric (1–5) to enable quantitative comparison. For each domain, we selected the top-K messages (K = 5, 10, 15, 20, 25) using three strategies: (1) LLM predicted ratings with Digital Twin prompting, (2) a human "oracle" using actual human ratings, and (3) random selection.

Code to reproduce our analysis is available at https://github.com/yiqunchen/LLM-smoking-cessation/. The source message data can be found at https://osf.io/4ux8q/overview.

**Results**

Figure 1 presents the overview of the study design. Detailed descriptive analyses of participant characteristics can be found in Hamoud et al (Hamoud et al., 2025). We display the best-performing configuration for each model across rating domains in Figure 2. Overall, personalized LLM-based digital twinshad the best performance. The best-performing personalized LLM pipelines achieved exact accuracies of 0.49 (content), 0.45 (coping), and 0.49 (quitting), and directional accuracies of 0.75, 0.66, and 0.70, respectively. These results outperformed both zero-/few-shot prompting (≤0.39) and supervised baselines (≤0.38). GPT-5's hybrid model achieved 0.47 accuracy (κ=0.25) for content, 0.43 (κ=0.24) for coping, and 0.45 (κ=0.27) for quitting. Grok-4-Fast's hybrid configuration attained the overall highest quitting accuracy (0.49, κ=0.30), while its digital twin achieved 0.49 (κ=0.24) for content and 0.45 (κ=0.24) for coping. Gemini-2.5-Pro's digital twin performed best for quitting (0.45, κ=0.26) and coping (0.43, κ=0.23). Digital twin personalization also elevated DeepSeek-R1 (0.44, κ=0.20 for content) despite modest zero-shot performance. Logistic regression and random forest baselines ranged from 0.30-0.38 accuracy with κ ≤0.12 across domains.

## 3.1 Comparison of modeling strategies

Zero-shot prompting across all LLMs produced 24.2-39.1% accuracy (mean 28-32% per model) with κ values near zero, only marginally exceeding supervised baselines. Incorporating six to ten examples in few-shot prompts yielded modest gains (≤4 percentage points). In contrast, digital twin personalization increased accuracy into the low-to-mid 40% range and raised κ to 0.18-0.24. Hybrid pipelines further improved performance for GPT-5 and Grok-4-Fast, adding 1-6 percentage points over the corresponding digital twin configurations. Performance heterogeneity across LLM vendors reached 10 percentage points within the same method family, reinforcing that model choice materially affects downstream accuracy.

Directional metrics highlight practical utility even when exact agreement is limited. Hybrid GPT-5 achieved 72.2% directional accuracy for content (Directional F1=0.56), 62.2% for coping (Directional F1=0.54), and 64.8% for quitting (Directional F1=0.55). Grok-4-Fast's hybrid quitting model yielded 70.0% directional accuracy (Directional F1=0.54). Macro-F1 scores remained modest (0.34-0.43), reflecting class imbalance and human variability, yet still exceeded supervised baselines by 8-12 points.

### 3.2 Distribution of predicted scores across models

Figure 3 shows the distribution of predicted scores across models. The true score distribution was concentrated in three categories, i.e., the one neutral and two more favorable categories, while the two negative categories were selected but less frequently. Traditional supervised learning models and zero-shot/few-shot LLMs tended to concentrate their predictions within one or two categories. In contrast, the digital twin models produced predictions that were more spread across all five categories.

### 3.3 Contrasting messages selected by LLM versus human raters

Figure 4 shows that LLM-selected messages, especially those chosen by Grok-4 and GPT-5, consistently achieved higher mean human ratings than random selection across all three domains. All LLMs fell behind the human oracle, but the gap was small (<0.5 on a five-point scale) and decreased as more messages were selected.

**Discussion**

In this study, we evaluated the capacity of LLMs to predict the perceived effectiveness of smoking cessation intervention messages, an essential step towards personalized digital health interventions. Across three message domains (content, coping, and quitting), zero-shot and few-shot LLMs performed similarly (best accuracies: 0.39, 0.34, and 0.33, respectively) to traditional machine learning baselines (accuracies: 0.30-0.38). Notably, personalized digital twin models substantially outperformed all other approaches (best accuracies: 0.49 for content, 0.45 for coping, and 0.49 for quitting). When evaluated using directional accuracy metrics, which assess whether models correctly predicted the relative ordering of message effectiveness, the best-performing digital twin models achieved directional accuracies of 0.75 (content), 0.66 (coping), and 0.70 (quitting). In addition to higher accuracy, digital twin predictions exhibited greater dispersion across rating categories compared with traditional ML and zero /few-shot LLMs, indicating improved sensitivity to individual-level heterogeneity in message perception.

Digital twin-based modeling presents substantial opportunities to advance the development of personalized mHealth interventions for smoking cessation. For example, it enables the personalized selection of previously tested messages, as well as the evaluation of new messages for each individual, based on the preferences learned from their digital twin. In addition to LLM-powered digital twins that are conditioned on static history like our study, they can also be conditioned on dynamically changing information (Makarov et al., 2025; Silva & Vale, 2025). As wearable sensors, smartphones, and passive monitoring technologies continue to generate rich streams of real-world behavioral and physiological data, records such as an individual's momentary behavior patterns, contextual states, and prior responses to treatment (Businelle et al., 2022; Hébert et al., 2025; Li et al., 2024; Lin et al., 2023; Luken et al., 2023; Thrul et al., 2025) could be valuable. Within smoking cessation, these data streams are particularly valuable given the pronounced heterogeneity in triggers, motivational trajectories, withdrawal symptoms, and relapse dynamics across individuals. By modeling these personalized patterns, digital twins may support the design of adaptive and context-aware intervention protocols, such as selecting coping messages most likely to be effective for a specific user at a specific moment or tailoring content to match dynamic risk states.

Our findings also highlight potential limitations of using models focusing on population level trends (e.g., traditional supervised learning and off-the-shelf zero-shot or few-shot LLMs) for identifying specific individual behavior, such as PME. Although prior work has mostly applied regression and supervised learning models to identify linguistic or contextual features associated with PME (Hamoud et al., 2025; Solnick et al., 2021; Tripp et al., 2021), in our study, traditional models' accuracy ranged from 30.3% to 37.6%. Similarly, despite growing evidence that zero-shot and few-shot LLMs (Brown et al., 2020; Li, 2023) can approximate human judgments, their performance in this task remained limited, achieving accuracies of only 0.33-0.39. In contrast, personalized digital twin models produced meaningful improvements, reaching 0.45-0.49 across the three message domains. While these gains highlight the promise of individual-level conditioning, the best-performing digital twins still leave room for improvement in fully predicting participants' exact ratings. One plausible explanation is the inherently noisy nature of individual human behavior. Even within-person consistency is imperfect: the test-retest reliability for individual choices in prior work has been estimated at approximately 81% (Park et al., 2024; Toubia et al., 2025). This suggests that a substantial proportion of variance in PME reflects intra-individual fluctuations, such as momentary affect, context, attention, or cognitive load, that are neither captured in static metadata nor easily recoverable from simple prompts.

Evaluated on directional consistency rather than exact category agreement, digital twin models achieved accuracies of 0.66–0.75, indicating that even when models missed the precise rating, they often correctly inferred positive, neutral, or negative attitudes. This distinction is meaningful because PME in our study relied on a 5-point scale, and prior work suggests that individuals may vary in how they apply such scales due to factors such as personality and culture (Kemmelmeier, 2016; Naemi et al., 2009; Pokropek et al., 2023). As a result, some individuals favor the endpoints of the scale, whereas others gravitate toward midpoints regardless of underlying judgments. Given this known variability in rating behavior, the practical value of exact-category prediction for personalized intervention design remains an open question. For many applications in digital health, particularly those aimed at selecting or tailoring message content, a model's capacity to correctly identify whether a message elicits a positive or negative reaction from an individual may be more consequential than matching the exact intensity of that reaction.

The distribution of predicted scores further helps contextualize why digital twin models outperformed both supervised learning and off-the-shelf zero-/few-shot LLMs. Digital twins' broader score dispersion suggests that they were more sensitive to capture nuanced differences in individuals' PME across messages, rather than defaulting to a narrow band of responses. Such behavior is consistent with the fact that digital twins incorporate person-specific historical characteristics and rating data, enabling them to learn individual preference patterns rather than population-level averages. For mHealth interventions focused on personalization, such as tailoring text-based smoking cessation interventions, this sensitivity is particularly valuable.

Despite their promising performance, several limitations of this study warrant consideration and highlight important directions for future work. First, this study draws on data from 301 young adult smokers who were members of an online market research panel, which may limit generalizability. Training digital twins on larger, more diverse samples would improve external validity and help determine whether the observed performance gains extend across demographic and behavioral subgroups. Second, the high noise of human ratings constrained the performance of the tested models. Future work may benefit from collecting multiple repeated assessments per message. Third, except for the models reported in this paper, we also explored incorporating pretrained language model embeddings to provide richer semantic representations of message content, these features did not meaningfully improve predictive accuracy. This suggests that static embedding-based semantic information alone may be insufficient for modeling fine-grained, individual-level differences in PME. Fourth, incorporating domain knowledge about message categories (e.g., CBT vs. ACT) produced only modest performance gains. This suggests that while domain information has potential value, its impact may be limited in static modeling settings. Future research could explore integrating time-series data within both LLM-based digital twins and other modeling frameworks, including transformer-based sequence architectures.Finally, while PME is a useful proximal indicator of intervention message quality, it may not fully capture actual behavioral impact (O'Keefe, 2018). Future studies should examine the application of LLM-based digital twin approaches in real-world intervention settings to determine their ability to optimize behavioral outcomes.

**Conclusion**

In conclusion, this study evaluated the efficacy of LLM-based approaches for predicting PME in the context of smoking cessation intervention messages. Digital twin models, which condition predictions on an individual's historical data, demonstrated the highest accuracy. These findings highlight the potential of LLM-based digital twins to support the development of more personalized and effective mobile health interventions for smoking cessation and other health behavior change targets. Further validation is needed in larger, more diverse populations and in prospective studies designed to evaluate real-world optimization and behavioral outcomes.

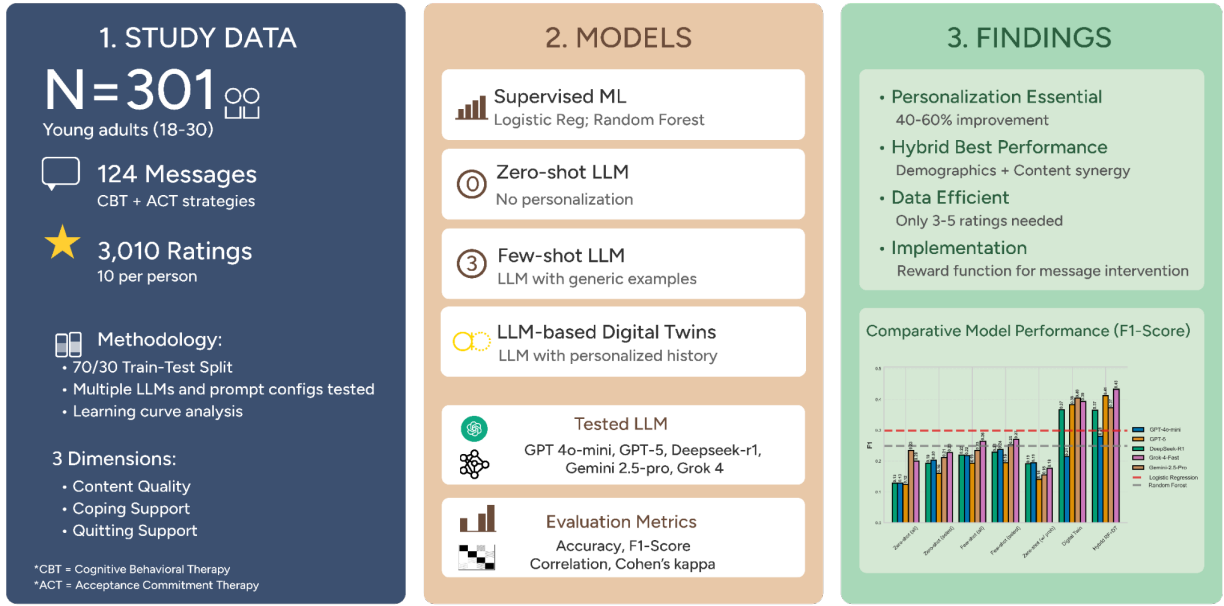

**Figure 1. Overview of study design, model types, and key findings in evaluating large language models for predicting self-rated smoking-cessation message effectiveness.** Panel 1 summarizes the study dataset of 301 young adults (ages 18–30) who rated 124 cognitive-behavioral and acceptance-commitment–based messages (3,010 total ratings). Panel 2 outlines the modeling approaches, including traditional supervised machine-learning baselines, zero-shot and few-shot large language models (LLMs), and personalized LLM-based "digital-twin" models. Panel 3 highlights our findings.

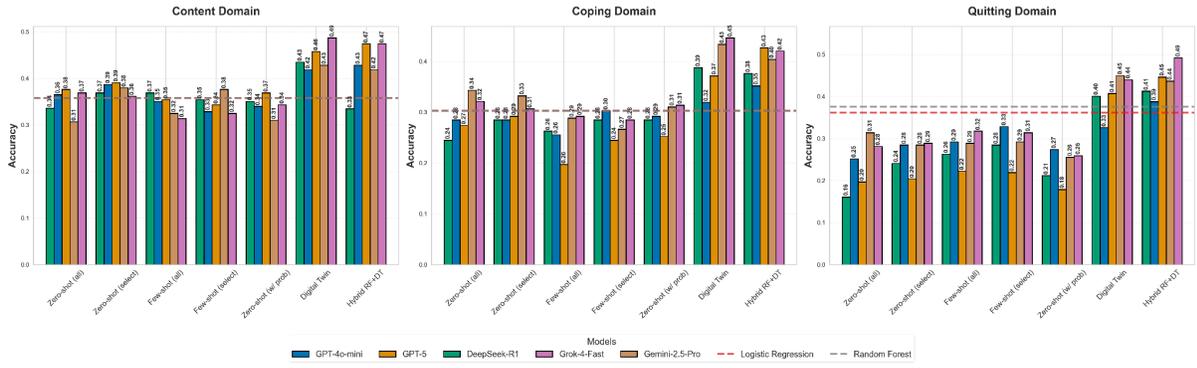

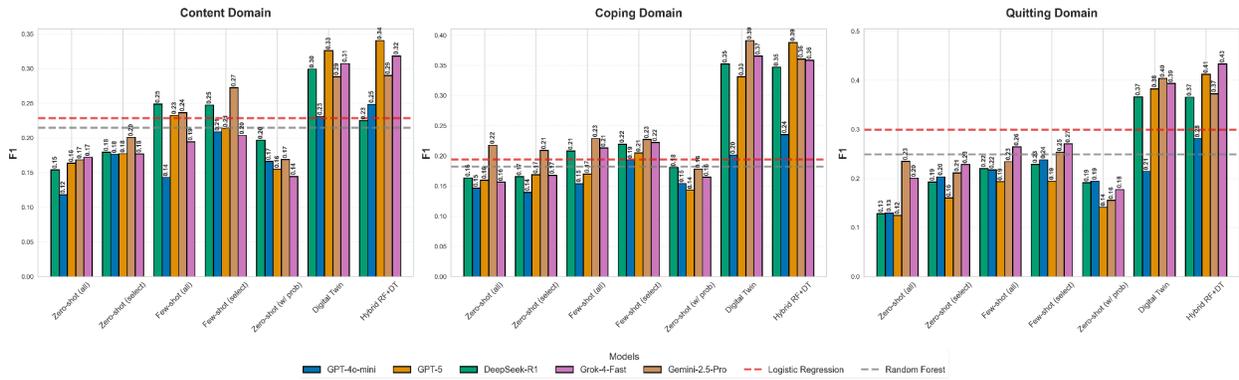

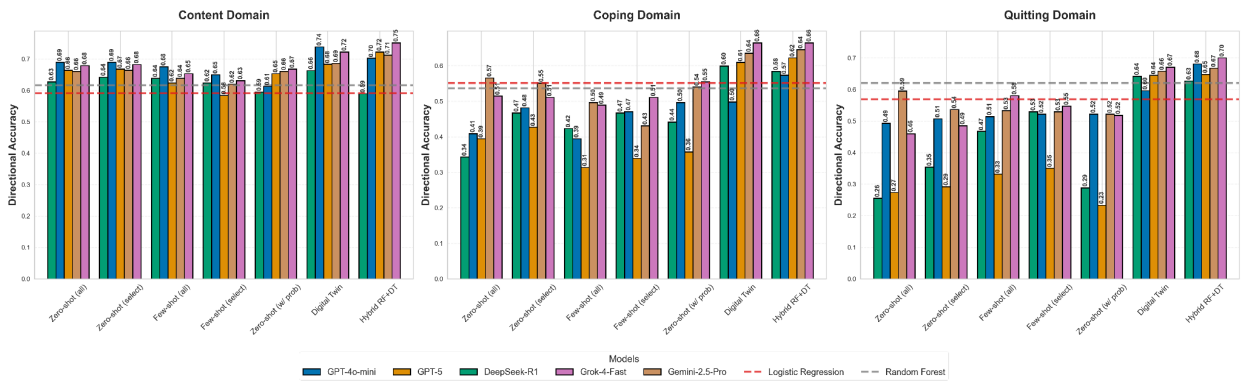

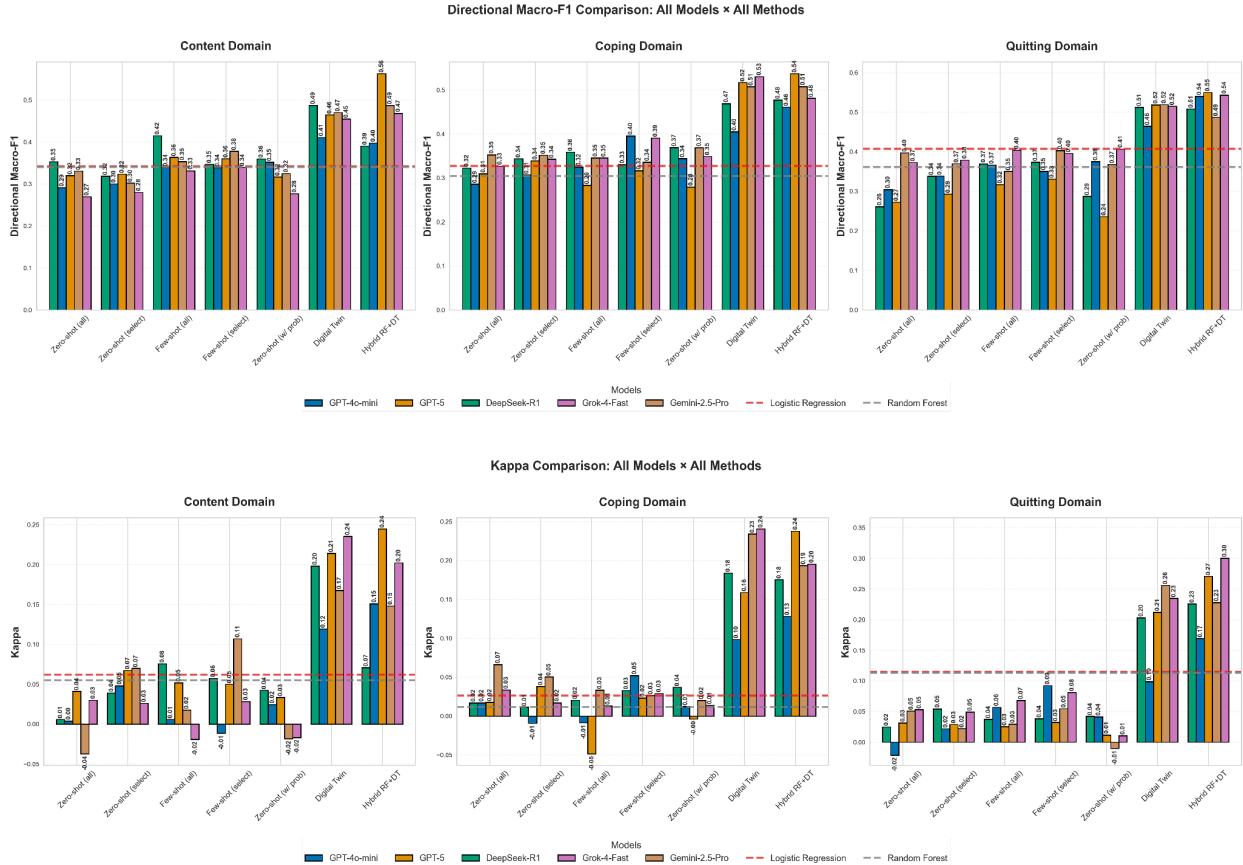

**Figure 2. Best-performing configuration for each model across rating domains.** Comparison of five evaluation metrics (accuracy, F1, directional accuracy, directional macro-F1, and Cohen's κ) across models (GPT-4o-mini, GPT-5, DeepSeek-R1, Grok-4-Fast, and Gemini-2.5-Pro) and configurations (zero-shot with all features, zero-shot with selected features, few-shot with all features, zero-shot with selected features and probability-like outputs, digital twin with all features, and digital twin with random forest results as priors) for predicting participant ratings in the Content, Coping, and Quitting domains. Each bar corresponds to a specific model–configuration combination. Dashed horizontal lines denote baseline performance from logistic regression (red) and random forest (gray).

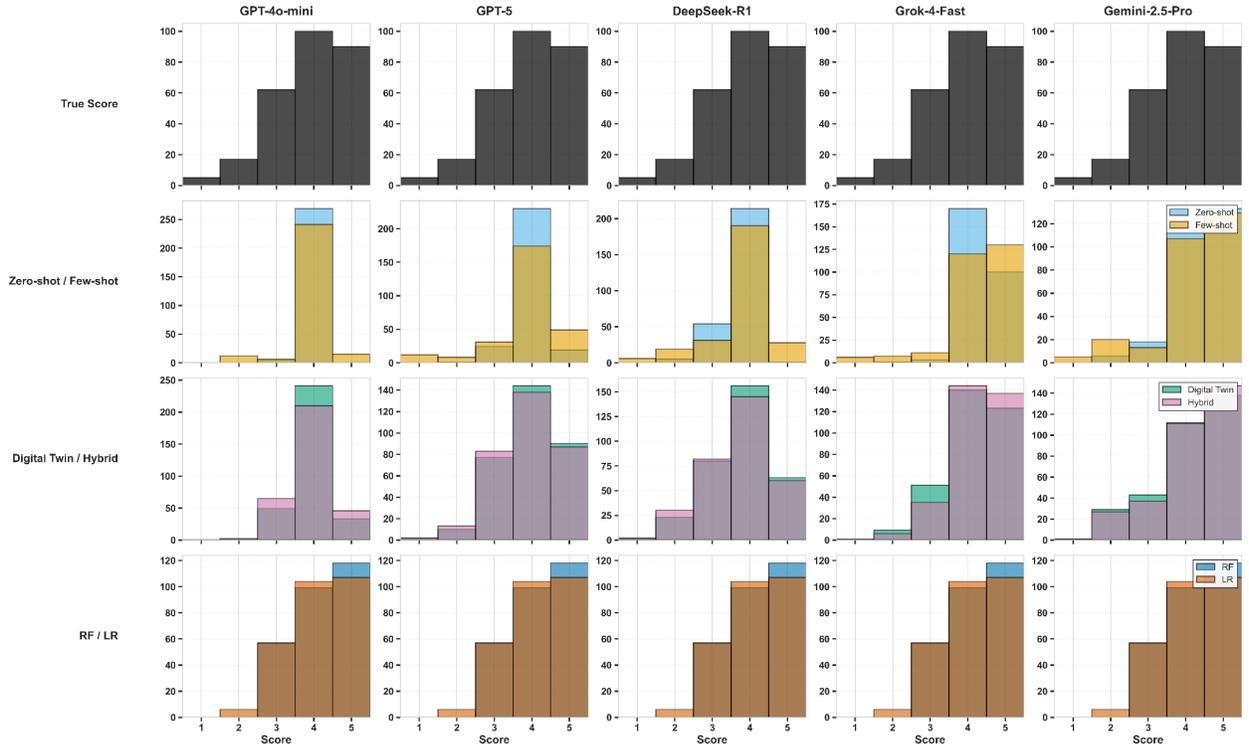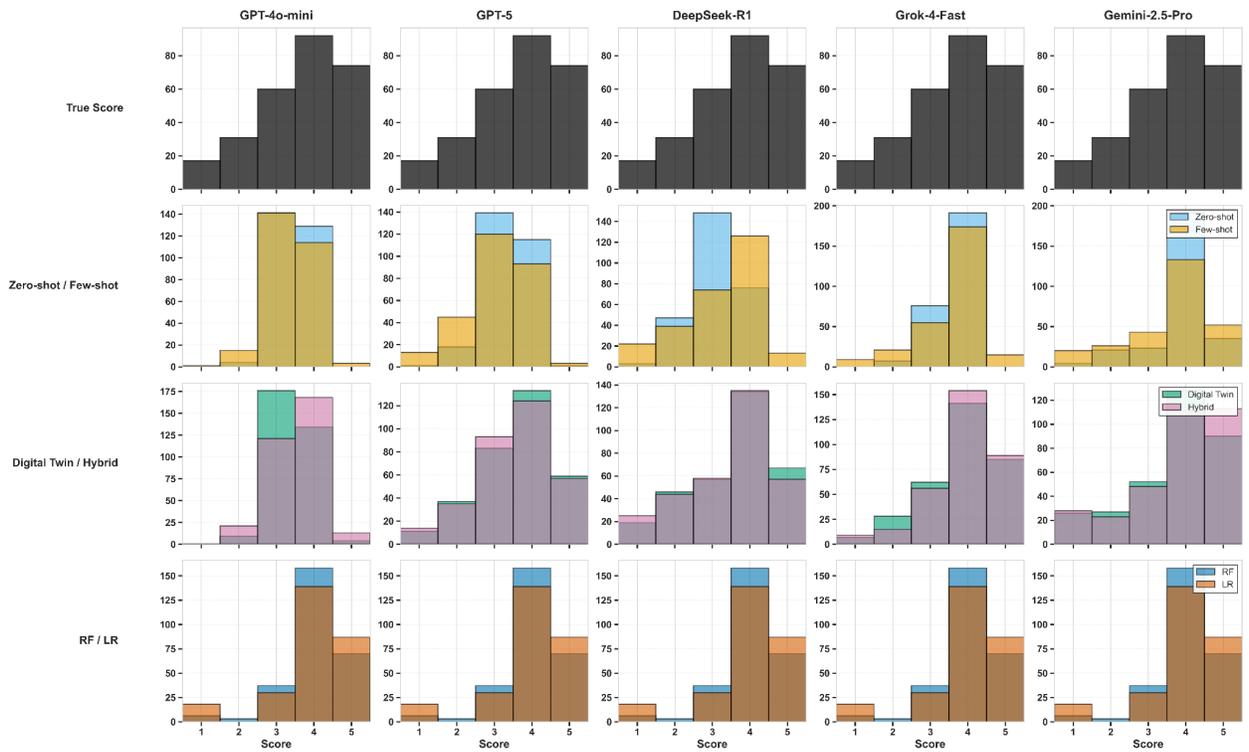

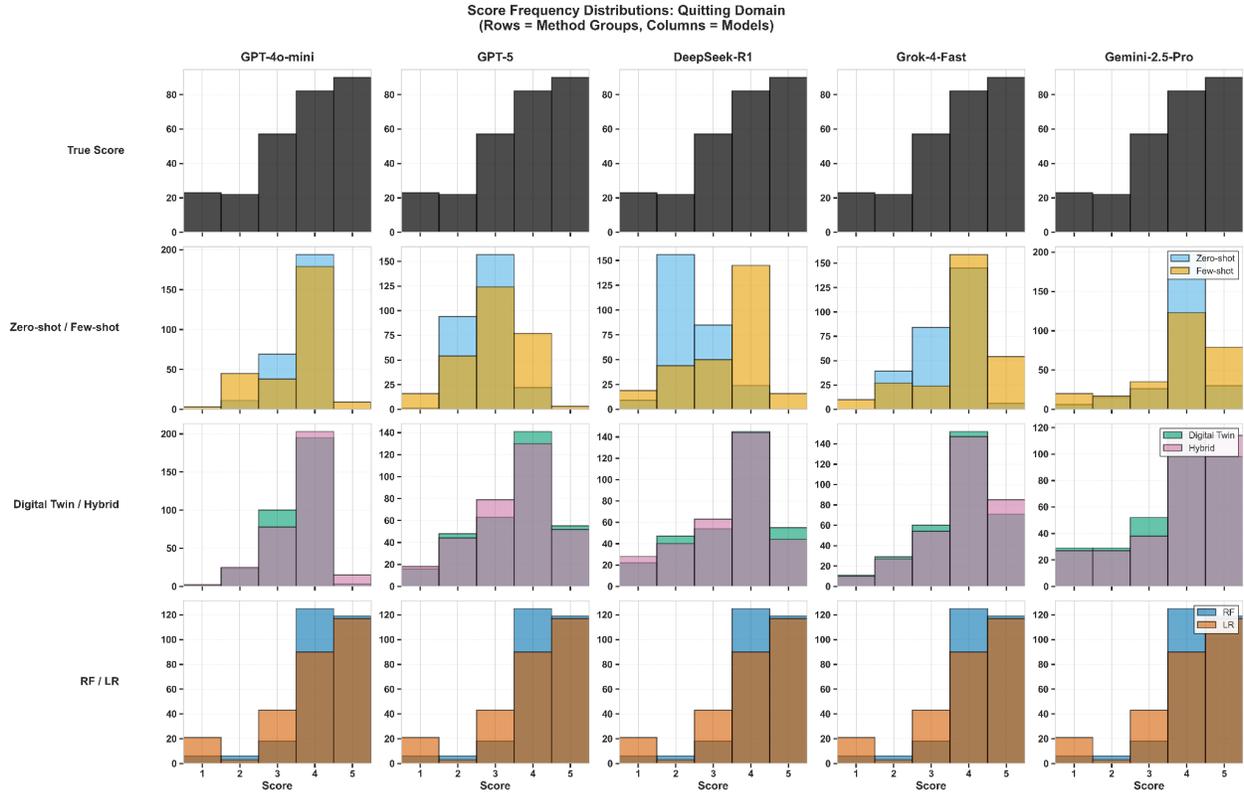

**Figure 3. Distribution of predicted scores across models.** Score frequency distributions in the content, coping and quitting domains across models and configurations. Histograms show the distribution of true participant scores (top row) and predicted scores from zero-shot/few-shot prompting, digital twin/hybrid approaches, and baseline machine-learning models (random forest and logistic regression). Columns correspond to language models.

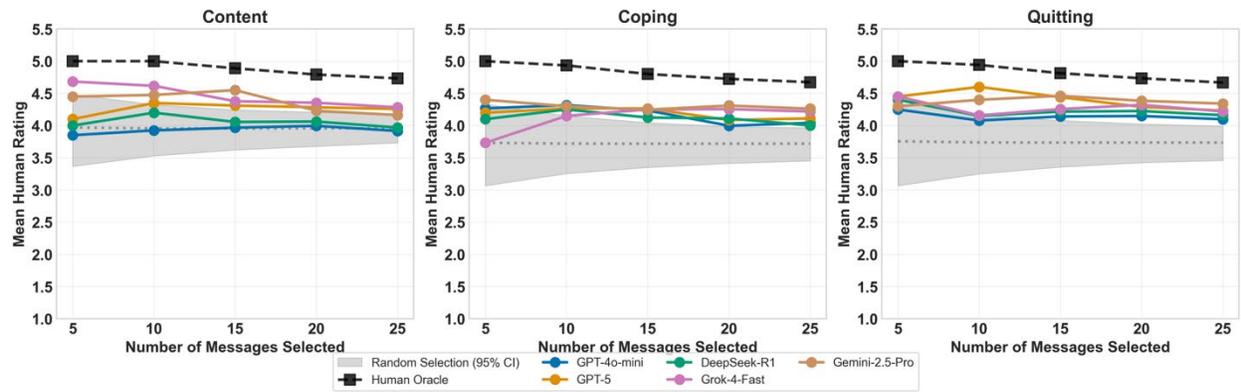

**Figure 4.** Mean human rating of top-K messages selected by LLMs (Digital Twin prompting), human oracle, and random selection across three evaluation domains. Shaded gray region indicates 95% CI for random selection. LLM-selected messages consistently outperform random selection and approach human oracle performance, particularly for larger K.

**Supplementary Materials for Personalized Prediction of Perceived Message Effectiveness Using Large Language Models**

**Appendix**

**Appendix A1: Prompt templates for LLMs.**

In this section, we provide the LLM prompt templates for the different configurations described in Sections 2.2.2 and 2.2.3.

Throughout the document:

- {{input_message}}, {{response_id}}, etc. are placeholders filled by the evaluation pipeline.

- {{metadata_block}} includes all participant features, and {{selected_metadata_block}} only includes age, gender identity, race/ethnicity, motivation to quit, and perceived social support.

- {{few_shot_examples_all}} and {{few_shot_examples_select}} summarize the pre-sampled examples (up to five examples).

- {{prob_metadata_block}} mirrors the selected-features list {{selected_metadata_block}} but retains the short labels used in the probability prompt.

### A1.1. Zero-shot with all features (Zero-shot (all))

**Description.** This template included all individual characteristics (sociodemographic variables, smoking history, and psychological flexibility items, totaling 23 features) and queried the LLMs to generate categorical predictions for content, design, coping, and quitting.

```
You are an expert in smoking-cessation communication and intervention.

Persona Setup
- Describe the provided image in one sentence to ground yourself visually.
- Then step into the shoes of the participant described below and judge a
 new message from their exact perspective.

Rating Dimensions
| Dimension | What it captures | Allowed ratings |
| content | Words + meaning quality | Very poor / Poor / Acceptable / Good / Very good |
| design | Visual presentation | Very poor / Poor / Acceptable / Good / Very good |
| coping | Helpfulness for in-the-moment urges | Not at all helpful / Som
```

```
ewhat helpful / Moderately helpful / Very helpful / Extremely helpful |
| quitting | Helpfulness for long-term quitting | Not at all helpful / So
mewhat helpful / Moderately helpful / Very helpful / Extremely helpful |

Decision Rules
- Stay consistent with the participant's motivation, nicotine dependence,
 environment, and expressed preferences.
- Favor extreme ratings when the message strongly fits or clashes with th
eir history; do not default to the middle.
- Keep explanations to two sentences per dimension, pointing to the most
relevant participant factors.

Inputs
Message to rate: "{{input_message}}"
Participant metadata (full set):
{{metadata_block}}

Required JSON Output
{
  "response_id": "{{response_id}}",
  "input_message": "{{escaped_input_message}}",
  "image_description": "One-sentence description of any provided image.",
  "predicted_content": "Very poor/Poor/Acceptable/Good/Very good",
  "predicted_design": "Very poor/Poor/Acceptable/Good/Very good",
  "predicted_coping": "Not at all helpful/Somewhat helpful/Moderately hel
pful/Very helpful/Extremely helpful",
  "predicted_quitting": "Not at all helpful/Somewhat helpful/Moderately h
elpful/Very helpful/Extremely helpful",
  "explanation": "<=2 sentences per dimension describing why this partici
pant would respond that way."
}
```

### A1.2. Zero-shot with selected features (Zero-shot (select))

**Description.** We restricted the metadata to five high-coverage variables associated with smoking patterns and cessation readiness: Age, gender identity, race/ethnicity, self-reported motivation to quit, and perceived social support. These variables were selected by correlation magnitude and used to test whether a compact yet informative set of characteristics could match the performance of the full template.

**Template delta.** Reuse the Zero-shot (all) instructions verbatim except replace the metadata block with:

```
Participant metadata (selected features only):
{{selected_metadata_block}}
```

### A1.3. Few-shot with all features (Few-shot (all))

**Description.** Extends Zero-shot (all) by prepending two exemplars with extreme ratings per domain ("Very good"/"Extremely helpful" and "Very poor"/"Not at all helpful") sampled from the training set. Each example includes participant characteristics and ratings so the model can contrast message content with profiles while seeing the full rating spectrum before answering.

```
Few-Shot Example Preamble
Here are participant-specific demonstrations showing how demographics inf
luence ratings. Treat them as ground-truth labels drawn from the training
 split.

{{few_shot_examples_all}}

Now, based on the examples above, evaluate the following message for the
given participant and output the JSON specified in Zero-shot (all).
```

### A1.4. Few-shot with selected features (Few-shot (select))

**Description.** Mirrors Few-shot (all) but truncates both the exemplar participant features and the target query to the five selected characteristics. This variant tests whether reference examples continue to improve performance when the prompt is aggressively compressed.

```
Few-Shot Example Preamble (Selected Features)
The following demonstrations show the full rating spectrum while only exp
osing the five selected participant features.

{{few_shot_examples_select}}

Now, based on the examples above, evaluate the following message for the
given participant using the Zero-shot (select) instructions and JSON cont
ract.
```

### A1.5. Continuous zero-shot with selected features (Zero-shot (w/ prob))

**Description.** Uses the same configuration as the continuous zero-shot with a natural-language profile but presents participant information in a concise bullet list rather than prose. Ratings are continuous (1.0–5.0) and accompanied by confidence scores; this variant isolates the effect of presentation format from information content.

```
You are an expert in smoking cessation communication. Predict how this pa
rticipant would rate the quality of a smoking-cessation support message b
y providing numerical scores between 1.0 and 5.0 for each dimension.

Instructions
1. Describe any provided image in one sentence.
```

```
2. Rate content, design, coping, and quitting on continuous 1.0–5.0 scale
s (decimals allowed) using the mappings:
   - 1.0 = Very poor / Not at all helpful, 5.0 = Very good / Extremely he
lpful.
3. Provide a confidence score (1.0–5.0) for each rating and keep the expl
anation to <=2 sentences total.

Input Message
"{{input_message}}"

Participant metadata (selected bullet list):
{{prob_metadata_block}}

Return JSON
{
  "response_id": "{{response_id}}",
  "input_message": "{{escaped_input_message}}",
  "image_description": "One-sentence description of any provided image.",
  "predicted_content": <float 1.0–5.0>,
  "predicted_design": <float 1.0–5.0>,
  "predicted_coping": <float 1.0–5.0>,
  "predicted_quitting": <float 1.0–5.0>,
  "confidence_content": <float 1.0–5.0>,
  "confidence_design": <float 1.0–5.0>,
  "confidence_coping": <float 1.0–5.0>,
  "confidence_quitting": <float 1.0–5.0>,
  "explanation": "<=2 sentences covering all four dimensions."
}
```

### A1.6 LLM-based Digital Twins

This framework extends the few-shot paradigm by contextualizing examples at the individual level, incorporating each participant's persona and entire historical responses to build participant profiles. Using these profiles, we evaluated two configurations.

**Basic profile (generate_digital_twin_prompt)**

**Description.** Combines the participant's full metadata with up to seven prior messages and their ratings. The model is instructed to simulate the participant, anchor predictions in message similarity, and stay faithful to historical responses.

```
Role & Task
You are an AI assistant simulating this participant. Predict how they wil
l rate a new message by comparing it to their metadata and previously rat
ed messages.

Ground Rules
- Stay faithful to past ratings and free-text feedback when available.
- Anchor your reasoning in similarities between the new message and the s
tored history.
```

```
- Keep coping/quitting judgments sensitive to message type whenever that
information is present.

Inputs
Message to rate: "{{input_message}}"
Participant metadata and history:
{{metadata_block}}
{{profile_messages_block}}

Output JSON Contract
{
  "response_id": "{{response_id}}",
  "predicted_content": "Very poor/Poor/Acceptable/Good/Very good",
  "predicted_design": "Very poor/Poor/Acceptable/Good/Very good",
  "predicted_coping": "Not at all helpful/Somewhat helpful/Moderately hel
pful/Very helpful/Extremely helpful",
  "predicted_quitting": "Not at all helpful/Somewhat helpful/Moderately h
elpful/Very helpful/Extremely helpful",
  "explanation": "<=2 sentences per dimension grounded in the participant
's profile and past reactions."
}
```

### Enhanced profile with RF priors (generate_hybrid_rf_digital_twin_prompt)

**Description.** Adds Random Forest predictions (trained on the full feature set) as auxiliary priors. The prompt inserts a section summarizing the RF outputs and instructs the LLM to treat them as one input among many, never as ground truth.

```
... (Basic profile template)

---
Additional Context: Prior Model Predictions
A Random Forest model trained on participant characteristics predicts:
- Content: {{rf_content_label}}
- Design: {{rf_design_label}}
- Coping: {{rf_coping_label}}
- Quitting: {{rf_quitting_label}}

These priors are not perfectly accurate. Use them alongside the participa
nt's history, the new message content, and your own judgment.

### OUTPUT FORMAT (reuse JSON contract)
```

**Appendix A2: Uncertainty quantification for LLM performance metrics.**

In this section, we present results from bootstrap analysis with n=1,000 resamples to quantify uncertainty in our performance estimates. For each bootstrap iteration, we resampled pairs of ground truth and predicted ratings with replacement and recomputed three metrics: (1) accuracy on the 5-point ordinal scale, (2) directional accuracy on a collapsed 3-point scale (low: 1–2, neutral: 3, high: 4–5), and (3) Cohen's kappa to assess agreement beyond chance. Confidence intervals are computed using the 2.5th and 97.5th percentiles of the bootstrap distributions (percentile method).

We listed the performance for digital twin (with CBT/ACT instructions); few shot (all features); and zero-shot (all features) in Tables S2-4. The bootstrap confidence intervals largely confirm the trends we summarized in the main text:

Across prompting strategies, the CIs reveal a clear hierarchy. Digital Twin prompting achieved statistically significant agreement (Cis for cohen's kappa excluding zero) in 15/15 model-domain combinations, compared to only 2/15 for Few-shot and 1/15 for Zero-shot.

Across models within a prompting strategy, Grok-4-Fast achieved the highest 5-class accuracy (43.7%–48.1% across domains) and most consistent directional accuracy (65.9%–71.1%), with CIs that consistently rank among the highest. GPT-4o-mini showed the lowest accuracy with CIs that occasionally overlap with but generally fall below other models.

Finally, across domains, the CIs reveal that Coping is consistently the most difficult domain across models—showing the lowest accuracy (32.8%–43.7%) and directional accuracy with CI generally ranked below other domains.

| Model | Domain | Accuracy [95% CI] | Dir. Acc. [95% CI] | Kappa [95% CI] |
|---|---|---|---|---|
| GPT-4o-mini | Content | 0.416 [0.363, 0.469] | 0.730 [0.680, 0.780] | 0.114 [0.048, 0.180] |
| GPT-4o-mini | Coping | 0.328 [0.279, 0.378] | 0.505 [0.449, 0.554] | 0.109 [0.059, 0.160] |
| GPT-4o-mini | Quitting | 0.325 [0.272, 0.378] | 0.598 [0.545, 0.647] | 0.095 [0.044, 0.150] |
| GPT-5 | Content | 0.453 [0.404, 0.509] | 0.674 [0.624, 0.724] | 0.208 [0.134, 0.285] |
| GPT-5 | Coping | 0.378 [0.325, 0.430] | 0.613 [0.554, 0.669] | 0.165 [0.099, 0.236] |
| GPT-5 | Quitting | 0.406 [0.356, 0.461] | 0.647 [0.598, 0.700] | 0.210 [0.145, 0.279] |
| DeepSeek-R1 | Content | 0.438 [0.385, 0.491] | 0.661 [0.612, 0.711] | 0.201 [0.129, 0.274] |

| Model | Domain | Accuracy [95% CI] | Dir. Acc. [95% CI] | Kappa [95% CI] |
|---|---|---|---|---|
| DeepSeek-R1 | Coping | 0.387 [0.331, 0.443] | 0.598 [0.542, 0.647] | 0.180 [0.105, 0.250] |
| DeepSeek-R1 | Quitting | 0.402 [0.359, 0.455] | 0.644 [0.594, 0.693] | 0.205 [0.147, 0.269] |
| Grok-4-Fast | Content | 0.481 [0.432, 0.540] | 0.711 [0.665, 0.758] | 0.230 [0.163, 0.308] |
| Grok-4-Fast | Coping | 0.437 [0.384, 0.492] | 0.659 [0.607, 0.715] | 0.226 [0.156, 0.297] |
| Grok-4-Fast | Quitting | 0.437 [0.384, 0.486] | 0.672 [0.622, 0.724] | 0.232 [0.161, 0.292] |
| Gemini-2.5-Pro | Content | 0.422 [0.366, 0.475] | 0.674 [0.621, 0.724] | 0.163 [0.085, 0.233] |
| Gemini-2.5-Pro | Coping | 0.433 [0.378, 0.489] | 0.635 [0.579, 0.687] | 0.233 [0.158, 0.302] |
| Gemini-2.5-Pro | Quitting | 0.449 [0.393, 0.502] | 0.659 [0.607, 0.709] | 0.256 [0.183, 0.327] |

Table S1. Bootstrap confidence intervals for Digital Twin method. Performance metrics include exact 5-class accuracy, directional 3-class accuracy and Cohen's kappa with 95% CIs calculated using the percentile method with n=1,000 message-level bootstrap resamples.

| Model | Domain | Accuracy [95% CI] | Dir. Acc. [95% CI] | Kappa [95% CI] |
|---|---|---|---|---|
| GPT-4o-mini | Content | 0.350 [0.299, 0.409] | 0.675 [0.624, 0.726] | 0.006 [-0.040, 0.052] |
| GPT-4o-mini | Coping | 0.255 [0.204, 0.307] | 0.394 [0.336, 0.453] | -0.009 [-0.068, 0.049] |
| GPT-4o-mini | Quitting | 0.292 [0.237, 0.347] | 0.515 [0.456, 0.569] | 0.057 [0.002, 0.115] |
| GPT-5 | Content | 0.354 [0.296, 0.409] | 0.624 [0.569, 0.682] | 0.052 [-0.018, 0.115] |
| GPT-5 | Coping | 0.197 [0.153, 0.252] | 0.314 [0.263, 0.369] | -0.049 [-0.105, 0.014] |
| GPT-5 | Quitting | 0.223 [0.175, 0.277] | 0.332 [0.270, 0.391] | 0.025 [-0.033, 0.082] |
| DeepSeek-R1 | Content | 0.369 [0.314, 0.423] | 0.639 [0.580, 0.697] | 0.076 [0.015, 0.145] |
| DeepSeek-R1 | Coping | 0.263 [0.215, 0.314] | 0.423 [0.369, 0.489] | 0.020 [-0.040, 0.083] |
| DeepSeek-R1 | Quitting | 0.263 [0.212, 0.318] | 0.467 [0.409, 0.526] | 0.037 [-0.023, 0.101] |
| Grok-4-Fast | Content | 0.314 [0.259, 0.365] | 0.653 [0.595, 0.712] | -0.019 [-0.091, 0.052] |
| Grok-4-Fast | Coping | 0.292 [0.234, 0.347] | 0.489 [0.427, 0.548] | 0.013 [-0.052, 0.078] |

| Model | Domain | Accuracy [95% CI] | Dir. Acc. [95% CI] | Kappa [95% CI] |
|---|---|---|---|---|
| Grok-4-Fast | Quitting | 0.318 [0.266, 0.369] | 0.580 [0.522, 0.639] | 0.068 [-0.001, 0.135] |
| Gemini-2.5-Pro | Content | 0.325 [0.270, 0.383] | 0.639 [0.577, 0.701] | 0.018 [-0.051, 0.094] |
| Gemini-2.5-Pro | Coping | 0.288 [0.237, 0.336] | 0.496 [0.438, 0.551] | 0.033 [-0.032, 0.093] |
| Gemini-2.5-Pro | Quitting | 0.288 [0.234, 0.343] | 0.533 [0.474, 0.591] | 0.029 [-0.042, 0.101] |

Table S2. Bootstrap confidence intervals for Few Shot (All) method. Performance metrics include exact 5-class accuracy, directional 3-class accuracy and Cohen's kappa with 95% CIs calculated using the percentile method with n=1,000 message-level bootstrap resamples.

| Model | Domain | Accuracy [95% CI] | Dir. Acc. [95% CI] | Kappa [95% CI] |
|---|---|---|---|---|
| GPT-4o-mini | Content | 0.365 [0.310, 0.420] | 0.690 [0.631, 0.741] | 0.004 [-0.017, 0.029] |
| GPT-4o-mini | Coping | 0.285 [0.234, 0.336] | 0.409 [0.354, 0.464] | 0.017 [-0.041, 0.075] |
| GPT-4o-mini | Quitting | 0.252 [0.201, 0.307] | 0.493 [0.431, 0.555] | -0.021 [-0.069, 0.031] |
| GPT-5 | Content | 0.376 [0.321, 0.431] | 0.664 [0.609, 0.719] | 0.041 [-0.005, 0.095] |
| GPT-5 | Coping | 0.274 [0.223, 0.325] | 0.394 [0.332, 0.456] | 0.018 [-0.042, 0.073] |
| GPT-5 | Quitting | 0.197 [0.153, 0.245] | 0.274 [0.226, 0.332] | 0.031 [-0.007, 0.071] |
| DeepSeek-R1 | Content | 0.336 [0.281, 0.391] | 0.628 [0.569, 0.679] | 0.006 [-0.052, 0.062] |
| DeepSeek-R1 | Coping | 0.245 [0.193, 0.296] | 0.343 [0.285, 0.398] | 0.017 [-0.043, 0.077] |
| DeepSeek-R1 | Quitting | 0.161 [0.120, 0.204] | 0.255 [0.204, 0.310] | 0.025 [-0.014, 0.065] |
| Grok-4-Fast | Content | 0.369 [0.314, 0.427] | 0.679 [0.624, 0.734] | 0.030 [-0.043, 0.107] |
| Grok-4-Fast | Coping | 0.321 [0.266, 0.376] | 0.515 [0.456, 0.573] | 0.033 [-0.022, 0.092] |
| Grok-4-Fast | Quitting | 0.281 [0.230, 0.332] | 0.460 [0.405, 0.518] | 0.053 [-0.010, 0.108] |
| Gemini-2.5-Pro | Content | 0.307 [0.259, 0.361] | 0.661 [0.609, 0.715] | -0.037 [-0.107, 0.040] |
| Gemini-2.5-Pro | Coping | 0.343 [0.288, 0.401] | 0.566 [0.511, 0.624] | 0.066 [0.008, 0.126] |
| Gemini-2.5-Pro | Quitting | 0.314 [0.255, 0.365] | 0.595 [0.540, 0.650] | 0.051 [-0.014, 0.108] |

Table S3. Bootstrap confidence intervals for Zero Shot (All) method. Performance metrics include exact 5-class accuracy, directional 3-class accuracy and Cohen's kappa with 95% CIs calculated using the percentile method with n=1,000 message-level bootstrap resamples.

### Appendix A3: Accuracy of digital-twin configurations by number of training ratings.

Supplementary Figure 1 illustrates the performance of the digital-twin prompt variants as a function of the number of historical ratings available for each participant (i.e., ratings of other messages used to condition the digital twin). Across models, accuracies remain remarkably stable even when using as few as one to seven prior message ratings. Accuracy exhibits substantially less variability than F1 or rank-based metrics such as Cohen's κ, which are more sensitive to class imbalance.

These preliminary findings suggest that a lightweight onboarding procedure, such as collecting only a small seed set of message-rating examples, may be sufficient for initializing reliable personalized digital-twin predictions before delivering tailored intervention content.

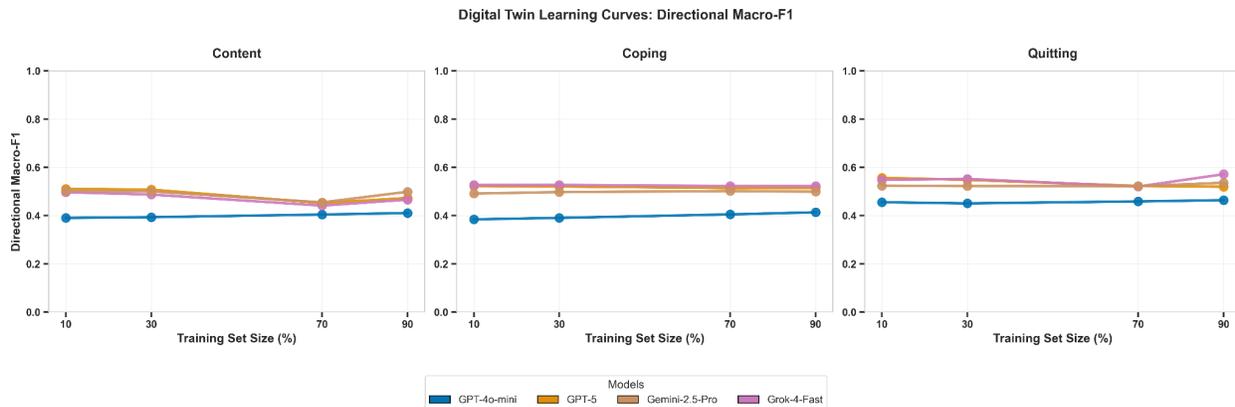

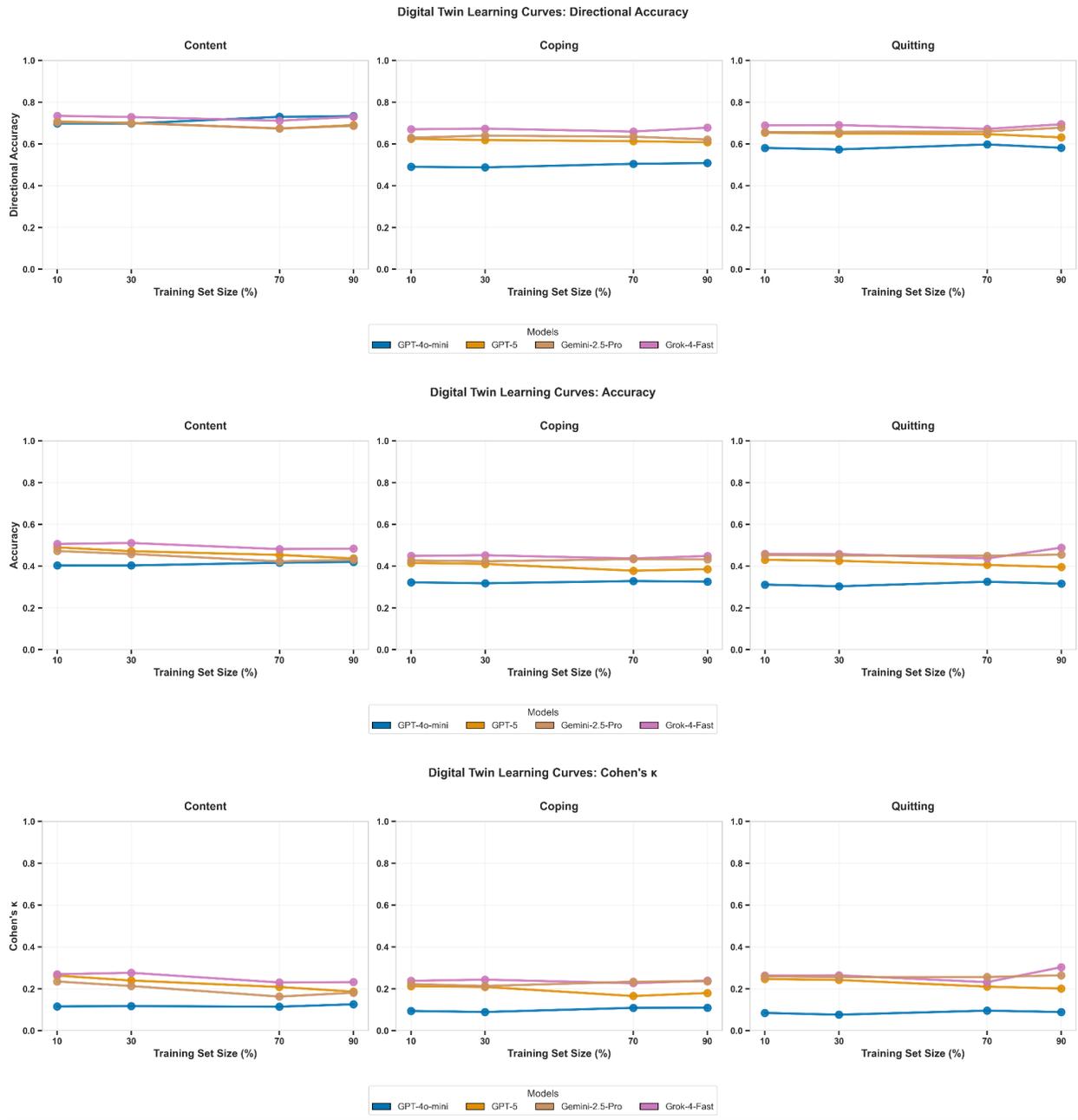

Figure 5. Model prediction accuracy for digital twin prompt configurations across training set sizes. Different color indicates different models.